\documentclass[11pt]{article}

\usepackage[final]{acl}
\usepackage{times}
\usepackage{fontawesome5}
\usepackage{latexsym}
\usepackage{amsmath} 
\usepackage{amssymb}
\usepackage[T1]{fontenc}

\usepackage[utf8]{inputenc}

\usepackage{microtype}

\usepackage{inconsolata}
\usepackage{hyperref}

\usepackage{graphicx}
\usepackage{inconsolata}
\usepackage{xcolor}%
\usepackage{textcomp}%
\usepackage{manyfoot}%
\usepackage{booktabs}%
\usepackage{algorithm}%
\usepackage{algorithmicx}%
\usepackage{algpseudocode}%
\usepackage{listings}%
\usepackage{colortbl}
\usepackage{booktabs}
\usepackage{float}
\usepackage{xcolor}
\usepackage{lineno}
\usepackage{booktabs}
\usepackage{array}
\usepackage{longtable}
\usepackage{pdflscape}

\usepackage{graphicx}
\definecolor{mygray}{gray}{0.8}
\definecolor{mygray2}{gray}{.95}
\definecolor{verdescuro}{rgb}{0.3,.7,0.3}
\definecolor{verdechiaro}{rgb}{0.6,.9,0.6}
\definecolor{lightblue}{rgb}{0.68, 0.85, 0.9}
\definecolor{lightindianred}{rgb}{0.87, 0.58, 0.58}

\newcommand{\val}[2]{#1\,\text{\fontsize{4.5}
{5}\selectfont$\pm$#2}}

\title{The Changing Geometry of Grammar: Dimensionality and Neighborhood Reorganization across Transformer Layers}

\newcommand*\samethanks[1][\value{footnote}]{\footnotemark[#1]}

\author{
Samuele Vallisa\thanks{The first two authors contributed equally to this work.}$^{\heartsuit}$, Federico Ravenda\samethanks$^{ \heartsuit,\displaystyle \star}$, 
\textbf{Claudio Palominos}$^{\heartsuit}$, 
\textbf{Rui He}$^{\heartsuit}$, \\
\textbf{Andrea Raballo}$^{\displaystyle \star,\clubsuit}$, 
\textbf{Antonietta Mira}$^{\displaystyle \star,\diamond}$, \textbf{Philipp Homan}$^{\triangle, \spadesuit}$, \textbf{Wolfram Hinzen}$^{ \heartsuit, \nabla}$\\
$^{\heartsuit}$ Grammar and Cognition Lab, Department of Translation \& Language Sciences, \\Universitat
Pompeu Fabra, Barcelona, Spain.\\$^{\displaystyle \star}$ Euler Institute, Università della Svizzera italiana, Lugano, Switzerland. \\
$^{\clubsuit}$ Cantonal Socio-Psychiatric Organization, Mendrisio, Switzerland. \\$^{\diamond}$ University of Insubria, Como, Italy.\\
$^{\triangle}$ Department of Adult Psychiatry and Psychotherapy, University Hospital of Psychiatry Zurich,\\
University of Zurich, Zurich, Switzerland.\\
$^{\spadesuit}$Neuroscience Center Zurich, University of Zurich and ETH Zurich, Zurich, Switzerland. \\
$^{\nabla}$ Institució Catalana de Recerca i Estudis Avançats, Barcelona, Spain.\\
\textbf{Corresponding authors:} \texttt{samuele.vallisa@upf.edu}, 
\texttt{federico.ravenda@usi.ch} 
\\[2cm]
}

\begin{document}
\maketitle

\begin{abstract}
Transformer representations describe trajectories through high-dimensional vector spaces, which are
shaped dynamically as tokens incorporate relational context across layers. Such data tend to concentrate on lower-dimensional sub-manifolds, a form of compression
quantified by the Intrinsic Dimensionality (ID), the minimum number of independent
variables needed to represent them without significant information loss.  
In this work, we ask whether the grammatical role of tokens, as marked by
their part-of-speech (PoS) tag, shapes the local geometry of this manifold. To this end:
\textbf{(1)} We investigate the layer-wise evolution of ID, finding that closed-class
items expand earlier and collapse sooner than open-class ones;
\textbf{(2)} We show its expansion and contraction to be explained by changes in the
neighborhood structure, and hence in the relations between words within a sentence;
\textbf{(3)} We compare encoders (ModernBERT, bigbird-roberta-large) and decoders
(gemma-2-2B, Llama-3.2-3B), finding that the two families evolve differently across
layers, consistently with how each integrates context;
\textbf{(4)} We show that geometric features alone recover a token's grammatical role, and use
them to interpret how the semantic content of each PoS evolves across layers in a
downstream classification task\footnote{\small \textcolor{darkblue}{\textbf{\small  Github Repository }}\href{https://github.com/Fede-stack/The-LLMs-Changing-Geometry-of-Grammar}{\faGithub}
}.

\end{abstract}

\section{Introduction \& Related Works}

Despite their widespread deployment, many
properties of transformer models remain poorly understood. A substantial body of work has
addressed this opacity through the geometry of internal representations, which, despite
living in high-dimensional spaces, concentrate on or near lower-dimensional manifolds
\citep{cai2021isotropy}, with the Intrinsic Dimension (ID) of activations relating to
information-theoretic compression \citep{cheng2023bridging}. The partition between
open-class and closed-class items is among the most robust in linguistic typology, and
the two classes differ precisely in where their content originates: \textit{content words} denote
lexically, \textit{function words} contribute relationally and are near/vacuous in isolation. If
contextualization is the mechanism by which a transformer resolves grammatical
dependencies, the two classes should not traverse the network alike. Two findings support
this: function words exhibit low self-similarity across contexts
\citep{ethayarajh2019contextual}, and ID reveals a high-dimensional abstraction phase in
decoders at the sequence level \citep{cheng2025emergence}. We bring the analysis to the
token level and condition it on parts-of-speech (PoS) distinctions, asking whether function and
content words trace distinct geometric trajectories through the network.
The main contributions of this work are:
\textbf{(1)} \textit{PoS-conditional} variants of two geometric descriptors: a new
closed-form conditional ID estimator for arbitrary token subsets, and a token-level
Information 
Imbalance (\textbf{II}) \cite{glielmo2022ranking} tracking how neighborhood structure is
reorganized across layers.
\textbf{(2)} Evidence that content and function words trace distinct trajectories, and
that contractions and expansions of the conditional ID trace back to identifiable
reorganizations of the local neighborhood.
\textbf{(3)} A comparison of encoders and decoders (\texttt{ModernBERT},
\texttt{bigbird}, \texttt{Gemma-2-2B}, \texttt{Llama-3.2-3B}) showing the coupling
between ID and neighborhood reorganization anchored to the last layer in the former and
to the input layer in the latter, consistently with their training objectives.
\textbf{(4)} A demonstration that these trajectories are \textit{predictive of
grammatical role}: a logistic regression on geometric features alone separates content
from function words and recovers fine-grained PoS categories, while layer-wise accuracy
on a downstream task tracks the ID and II profiles, tying gains in semantic
informativeness to identifiable geometric reorganizations.

\section{Methods}

\noindent\textbf{Research Questions}. Our work is shaped by the following research questions:\\
\noindent\textbf{(RQ1)} Does the geometry of a word's representation reflect its
lexical class? Do open-class and closed-class items, which differ in whether their
content is lexical or relational, exhibit systematic geometric differences?\\
\noindent\textbf{(RQ2)} How do neighborhoods reorganize across layers, and does resolving a dependency correspond
to a contraction or expansion of the ID?\\
\noindent\textbf{(RQ3)} Does the direction in which context is integrated, bidirectional
versus causal (i.e., encoder vs. decoder), alter these trajectories?

\subsection{Dataset \& Models used}
To explore this idea, we used the \texttt{Pile-10k} dataset \citep{pile}\footnote{\href{https://huggingface.co/datasets/NeelNanda/pile-10k}{\textcolor{blue}{https://huggingface.co/datasets/NeelNanda/pile-10k}}}, a text corpus comprising samples from 22 different sources. To obtain stable ID estimates, we retained only documents between 1'000 and 1'500 words, yielding \textbf{539} samples (on average, each sentence yields \textit{1'218} word representations). Beyond its use for fine-tuning, this dataset has been widely adopted in the literature to study the internal geometry of LLM representations \citep{viswanathan2025probing,viswanathan2025geometry,nielsen-etal-2025-prediction,kulkarni2026disentangling}.
We analyze two encoder (\texttt{ModernBERT}, \texttt{bigbird-roberta-large}) and two decoder (\texttt{Gemma-2-2B}, \texttt{Llama-3.2-3B}) models. The two encoders were selected because they support context lengths  >4'000 tokens, so that retained documents  fit without truncation.
Each token in the selected corpus was assigned a PoS tag using 
\texttt{en\_core\_web\_sm}\footnote{\href{https://spacy.io/models/en\#en_core_web_sm}{\textcolor{blue}{https://spacy.io/models/en\#en\_core\_web\_sm}}}, one of the pre-trained models included in the SpaCy library \cite{vasiliev2020natural}.
Among the set of PoS tags obtained, we filtered out special tokens (e.g., padding markers, end-of-text indicators) and aligned
SpaCy's tokenization with that of each model to prevent mismatches.
We restrict our analysis to the 8 most frequent PoS categories (excluding punctuation),
accounting for 83\% of PoS occurrences in our dataset. These comprise the four major
open classes (\textbf{NOUN}, \textbf{VERB}, \textbf{ADJ}, \textbf{ADV}) together with \textbf{PROPN} and the principal closed classes
(\textbf{DET}, \textbf{PRON}, \textbf{ADP}), and thus cover both poles of the lexical-functional distinction the
analysis targets; the excluded categories are either marginal in frequency or
heterogeneous in status (e.g.\ SYM, X) \cite{nivre2020universal}.

\subsection{Geometric Descriptors}

We characterize the local geometry of the representation manifold through two
complementary PoS-conditional measures, both computed within a single document and then
averaged across documents.

\noindent\textbf{PoS-conditional intrinsic dimension (cID).}
Although token representations $\mathbf{h}_i$ live in an ambient space of dimension $D$,
they are supported on a manifold of lower dimension,
$\{\mathbf{h}_i\}_{i=1}^{N} \subset \mathcal{M} \subset \mathbb{R}^{D}$ with
$d = \dim(\mathcal{M}) \ll D$, so that $d$ measures the effective degrees of freedom of
the representations. Read linguistically, $d$ estimated over a single PoS quantifies how
many independent factors condition the representation of that category: a low value
indicates items whose contextual variation is constrained, while a high one indicates items free to vary
along many dimensions at once. We estimate it by adapting ABIDE \citep{di2024beyond}, a recent
nearest-neighbor-based ID estimator robust to noise and to the choice of scale. Because the estimator depends only on the local geometry around each point and not on how
densely that region is populated, its likelihood factorizes across points, so restricting
it to the tokens of a given PoS yields a closed-form conditional estimator describing how
that category sits within the shared manifold. Full derivation in Appendix~\ref{posid}.
\begin{figure*}[t]
    \centering
    \includegraphics[width=1.\linewidth]{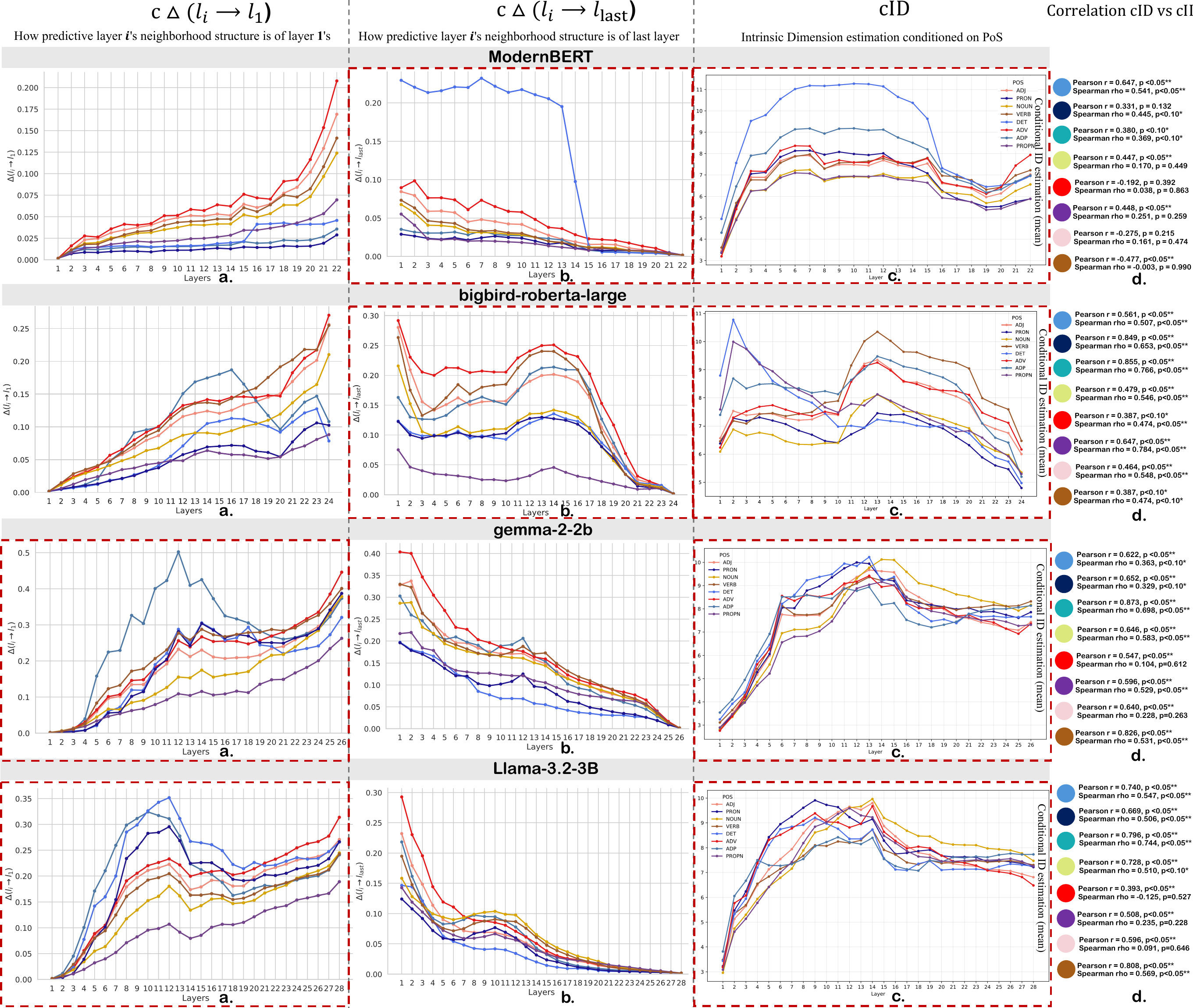}
    \caption{PoS-conditional II, averaged over documents, from layer $\ell$ to
layer 1 \textbf{(a.)} and to the last layer \textbf{(b.)}, for two encoders
(\texttt{ModernBERT}, \texttt{bigbird}) and two decoders (\texttt{gemma},
\texttt{Llama}). Panel \textbf{(c.)} shows the PoS-conditional ID across layers, and
\textbf{(d.)} the Pearson and Spearman correlations between the cID and cII trajectories.}
    \label{all_in_all}
\end{figure*}

\noindent\textbf{PoS-conditional information imbalance (cII).}
While cID quantifies how compressed the manifold is, it is blind to the organization of
the local neighborhoods: the same ID can correspond to very different geometries.
Information Imbalance \citep{glielmo2022ranking} fills this gap by measuring how well the
neighborhood structure of one layer predicts that of another, with low values indicating
preserved neighborhoods and high values independent ones. Linguistically, this tracks
when a token's company changes: a low $\Delta_p(\ell \to \ell_{\text{first}})$ means
tokens of category $p$ keep the distributional neighbors they had at the input, whereas a
rise signals reattachment to different items, presumably the constituents they enter into
a grammatical relation with. We compute it at the token level within a single sentence,
conditioning only through the averaging set: nearest neighbors and ranks are drawn from
all tokens, so that cross-category reattachment (e.g.\ a determiner detaching from other
determiners and moving toward its head noun) remains visible. Computing
$\Delta_p(\ell \to \ell_{\text{first}})$ and $\Delta_p(\ell \to \ell_{\text{last}})$ for
every layer localizes where a category detaches from its input identity and converges
toward its final contextual role. Formal derivation in Appendix~\ref{posii}.

\section{Results}

\noindent\textbf{Layer-wise geometry across PoS (RQ1, RQ2).}
Figure~\ref{all_in_all} reports, for each model: \textbf{(a)} cII from layer $\ell$ to layer 1; \textbf{(b)} cII from $\ell$ to the last layer;
\textbf{(c)} cID trajectories per PoS (lower values indicate preserved neighborhoods); \textbf{(d)} Pearson and Spearman correlations between the two. In
\texttt{ModernBERT}, function words exhibit a broad cID plateau in the middle layers,
preceding the peak of content words (normalized trajectories Figure~\ref{nonormalized} in
Appendix). DET contracts sharply at layer 16, coinciding with the
abrupt change in neighborhood structure in panel (a), and a second negative peak at layer
19 anticipates a reorganization affecting all PoS. In \texttt{bigbird-roberta}, DET, PRON and
PROPN instead follow a separate, decreasing trajectory over the first half of the network
before converging on a shared cID peak at layer 13, mirrored in panel (b); we attribute
this divergence to its very different training strategies.
 Decoders are more
homogeneous: in \texttt{gemma}, function words with ADV and VERB peak at layer 6 and
again at 11-13, whereas NOUN and ADJ peak later; \texttt{Llama} peaks at layers 9 (PRON,
ADV, DET), 15 (all PoS) and 19 (PRON, DET), with ADP earliest, at layer 4.\\
\noindent\textbf{Relating cII and cID (RQ2, RQ3).}
For encoders we correlate cID with cII($\ell \rightarrow$ last): bidirectional attention
mixes context aggressively in the first layers, so cII($\ell \rightarrow 1$) saturates
immediately and carries little variance across depth, whereas cII($\ell \rightarrow$
last) tracks the process cID describes, the convergence of each token toward the final
contextual role a masked-language objective is trained to recover. Correlations are
frequently significant, and the categories coupled in one encoder, predominantly
\textit{function} words, remain so in the other. For decoders we correlate against
cII($\ell \rightarrow 1$): causal attention integrates context incrementally, so
input-level geometry decays gradually and retains a wide dynamic range, while the last
layer collapses onto the unembedding space, encoding the \emph{upcoming} token rather
than the contextual identity of the current one, and is thus decoupled from the
abstraction phase cID captures.\\
\noindent\textbf{A shared trend.}
Across all four models the spread of cID between PoS narrows in the last layers,
suggesting progressive alignment driven by the grammatical relations established: categories differ not in their endpoint but in the path taken to reach it,
with divergence concentrated in the early and middle layers. Appendix~\ref{classtest}
tests whether this alignment is layer-dependent, tracking the accuracy of a classifier
fitted on the average encoder embedding of each PoS and lexical class. We find a
significant negative association between layer-mean cID and performance, strongest for
DET, PRON and ADP, with gains localized precisely where neighborhood structure is
reorganized: the layers at which grammatical role is most linearly decodable are those
where the local manifold is most compressed.\\
\noindent\textbf{The special status of function words (RQ1, RQ2, RQ3).}
This behavior is consistent with the status of ADP, PRON and DET as closed-class items
carrying little lexical content: their contribution is relational rather than
denotational, so the model first expands their representations to accommodate the
structural configurations they can enter, then collapses them once the dependency is
resolved. The cII profiles reveal an architectural dissociation on the same categories.
In encoders, function words remain the most predictive of the first layer, indicating
little neighborhood restructuring; in decoders the pattern reverses, with the weakest
predictivity of the input layer in the intermediate layers. Bidirectional attention gives
a determiner or an adposition immediate access to its complement, so the dependency is
available essentially from the input; under causal attention the governed element lies to
the right of the governor and the relation can only be established once the complement
has been processed, precisely where the restructuring is observed. PROPN patterns differently,
showing consistently low cII: as rigid designators, proper nouns denote relatively independently of
the predicates they combine with, and their neighborhood structure is correspondingly
stable throughout.

\begin{table}[t]
    \centering
    \tiny
    \setlength{\tabcolsep}{4pt}
    \begin{tabular}{lccccc}
    \hline
    
        \textbf{Method} & \textbf{Accuracy} &  \textbf{Precision} &\textbf{Recall} &  \textbf{F1} & \textbf{ROC-AUC} \\
        \hline
        \multicolumn{6}{c}{\cellcolor{mygray} \texttt{content vs. function words}} \\ \hline
        ID + II & \val{0.878}{0.008} & \val{0.861}{0.012} & \val{0.802}{0.017} & \val{0.830}{0.012} & \val{0.937}{0.005} \\ \hline
        II & \val{0.830}{0.007} & \val{0.836}{0.014} & \val{0.677}{0.016} & \val{0.748}{0.012} & \val{0.886}{0.006} \\ \hline
        ID & \val{0.850}{0.007} & \val{0.815}{0.014} & \val{0.773}{0.015} & \val{0.793}{0.009} & \val{0.907}{0.005} \\ \hline
        Shuffled & \val{0.593}{0.013} & \val{0.441}{0.019} & \val{0.349}{0.025} & \val{0.389}{0.021} & \val{0.618}{0.014} \\ \hline
        \multicolumn{6}{c}{\cellcolor{mygray} \texttt{PoS Classification}} \\ \hline
ID + II & \val{0.529}{0.010} & \val{0.567}{0.010} & \val{0.529}{0.010} & \val{0.540}{0.010} & \val{0.844}{0.005} \\ \hline
        II & \val{0.424}{0.008} & \val{0.486}{0.009} & \val{0.424}{0.008} & \val{0.441}{0.009} & \val{0.789}{0.006} \\ \hline
        ID & \val{0.486}{0.012} & \val{0.512}{0.013} & \val{0.486}{0.012} & \val{0.487}{0.013} & \val{0.816}{0.006} \\ \hline
        Shuffled & \val{0.171}{0.006} & \val{0.207}{0.011} & \val{0.171}{0.006} & \val{0.167}{0.007} & \val{0.534}{0.007} \\ \hline
            
    \end{tabular}
        \caption{Binary classification (\textbf{function} words vs.\ \textbf{content} words) and PoS classification performance.}
    \label{table1}
\end{table}

\subsection{Discriminative capacity of token geometric evolution}

To assess whether the layer-wise evolution of cID and cII captures syntactic properties of
tokens, we train logistic regression classifiers \emph{on geometric features alone}, with
no lexical or contextual input, on two tasks: \textbf{(1)} a \textbf{binary} task
predicting whether a token is a function word (DET, PRON, ADP) or a content word (VERB,
NOUN, ADV, ADJ), and \textbf{(2)} a \textbf{multiclass} task predicting its specific PoS
over the same seven categories. PROPN is excluded from both, as it cuts across the partition tested (consistent with its cID and cII trajectories): an open class that
nonetheless denotes rigidly \citep{kripke1980naming} and, unlike common nouns, is
argumental without a determiner \citep{longobardi1994reference}. Results including
PROPN are reported in Appendix~\ref{pointwise}. We sample $10'000$
random tokens from \texttt{Pile-10k} documents of $500$-$1'000$ words, with an $80/20$
train-test split, and extract point-wise ID and II trajectories for each token and model
(details in Appendix~\ref{pointwise}). These serve as feature vectors across four configurations: ID + II, ID only, II only, and
a Shuffled control permuting each token's features. We report accuracy, precision, recall,
macro-F1 and ROC-AUC over $20$ repetitions of the pipeline.
Geometric features alone discriminate content from function words
(F1 $=0.878$) and recover individual PoS categories (F1 $=0.540$), well above the
shuffled baselines ($0.389$ and $0.167$). Table~\ref{table1} reports the aggregate
results combining features from all four models; per-model results are in Appendix
Tables~\ref{binary_comparison},\ref{pos_comparison}. This confirms that the layer-wise
geometry of a token encodes its grammatical role, with no access to lexical identity.
\section{Conclusions}

PoS-conditional ID and II show that a token's geometric trajectory reflects its
grammatical status: closed-class items expand into higher-dimensional neighborhoods
earlier and collapse sooner than open-class ones, tracking the resolution of the elements
they combine with. The process is architecture-dependent, tracking whether attention
grants access to those elements from the input layer or only once encoded. That geometric
trajectories alone recover grammatical role further confirms that this role is encoded in the local geometry
of the manifold, independently of lexical identity.

\section{Limitations}

Our analysis is restricted to English. Since the geometric signatures we identify are
tied to grammatical properties that vary typologically, such as head directionality,
which underlies our account of the encoder/decoder dissociation, and the extent to which
grammatical relations are marked by free function words rather than by morphology,
languages with different typological profiles may well exhibit different trajectories.
Extending the analysis cross-linguistically is a natural direction for future work.

A second limitation concerns model coverage. We examine two encoders and two decoders,
which cannot be taken as representative of the full space of transformer language models:
architectural choices, pretraining data, model scale and training objective may all
influence the geometry of the resulting representations, and different models may
therefore yield different patterns. That said, the findings we report are consistent
across the four models considered, and the classification experiments show that the
geometric features remain predictive of grammatical role in every case, suggesting that
the phenomena we describe are not artifacts of a particular architecture. Verifying
whether they persist at larger scales and across a broader range of models remains open.

\bibliography{custom}

\appendix

\section{Derivation of the Geometric Descriptors }
\subsection{POS-Conditional Intrinsic Dimension Estimation}\label{posid}

The idea of ID is to quantify the complexity of high-dimensional datasets. In essence, it represents the minimum number of variables needed to describe the underlying structure of data without significant loss of information \cite{denti2022generalized}.

To estimate the ID of the representations conditioned on a specific PoS category, we adapt the Adaptive Binomial Intrinsic Dimension Estimator (ABIDE) \citep{di2024beyond}. ABIDE models the number of neighbors falling within two concentric balls centered at each
point. Given radii $r_B$ and $r_A = \tau r_B$ with $\tau \in (0,1)$, and denoting by
$k_{A,i}, k_{B,i}$ the neighbor counts in the two balls around point $i$, ABIDE exploits
the fact that, under local density homogeneity,
$$
k_{A,i} \mid k_{B,i} \sim \mathrm{Binomial}\big(k_{B,i},\, \tau^{d}\big),
$$
where $d$ is the intrinsic dimension of the manifold on which the points lie, the
quantity we want to estimate. It enters the model through the volume ratio of the two
balls: on a $d$-dimensional manifold the volume of a ball scales as $r^{d}$, so
$\mu(A)/\mu(B) = \tau^{d}$, and a neighbor already inside $B$ falls into $A$ with exactly
that probability. The local density $\rho_i$ cancels in this ratio, so each per-point
factor depends only on the local geometry and not on the density regime of the point.
Intuitively, $d$ is recovered by observing how fast the neighbor count shrinks when the
radius is reduced by a factor $\tau$: the faster the drop, the higher the dimension.

This property makes the likelihood separable across points, so its restriction to a subset $G$ (the tokens of a given PoS) is a legitimate conditional likelihood for $d$:
$$
L_G(d) = \prod_{i \in G} \binom{k_{B,i}}{k_{A,i}} (\tau^{d})^{k_{A,i}} (1-\tau^{d})^{k_{B,i}-k_{A,i}}
$$
whose maximizer has closed form
$$
\hat d_G = \frac{\log\big(\overline{k_A}^{\,G} / \overline{k_B}^{\,G}\big)}{\log \tau},
\qquad
\overline{k_A}^{\,G} = \frac{1}{|G|}\sum_{i \in G} k_{A,i}
$$
Neighbor counts include tokens of any PoS: the binomial conditional holds regardless of the labels of the points falling in the balls, provided density is locally uniform.

In practice, for each (document, layer) pair we run the full iterative ABIDE procedure on \emph{all} tokens, the optimal per-point neighborhood sizes $k^*_i$ is determined globally, and only restrict the final average to each PoS group. Uncertainty per group is obtained from the restricted Fisher information (Cramér–Rao bound with $|G|$ in place of $n$); groups with fewer than 10 tokens are discarded, to respect the asymptotic guarantees of the estimator.

\subsection{POS-Conditional Information Imbalance Across Layers}\label{posii}

To interpret the intrinsic dimension estimates, we complement them with a finer-grained analysis of how the representation geometry evolves layer by layer and how neighborhood structure differs across parts of speech. To this end, we quantify how the neighborhood structure of a token's contextual representation evolves across layers, and whether this evolution differs systematically by part of speech, by introducing a POS-conditional variant of the \emph{information imbalance}  \citep{glielmo2022ranking}, previously used by \citet{cheng2025emergence} to characterize a high-dimensional abstraction phase in language models at the sequence level. We adapt the measure along two dimensions: we compute it at the token level within a
single document, rather than over sequence-level representations, and we condition it on
PoS by restricting the average to the tokens of a given category.

For an input of $N$ tokens, let $\mathbf{h}^{\ell}_i \in \mathbb{R}^D$ denote the representation of token $i$ at layer $\ell$, and let $p_i$ denote its PoS tag. For two layers $A, B$, define the nearest neighbor of token $i$ in layer $A$'s representation space as
$$
j^A(i) = \arg\min_{j \neq i} \, d\big(\mathbf{h}^A_i, \mathbf{h}^A_j\big)
$$
and the rank of an arbitrary token $j$ relative to $i$ in layer $B$'s space as
\small
$$
r^B(i,j) = 1 + \big|\{m \neq i,j : d(\mathbf{h}^B_i,\mathbf{h}^B_m) < d(\mathbf{h}^B_i,\mathbf{h}^B_j)\}\big|
$$
\normalsize
Both quantities are computed over \textbf{all} $N$ tokens in the sentence, irrespective of part of speech.

Let $S_p = \{i : p_i = p\}$ be the set of token indices with POS tag $p$. We define
$$
\Delta_p(A \to B) = \frac{1}{|S_p|} \sum_{i \in S_p} r^B\big(i, j^A(i)\big) \Big/ \frac{N}{2}
$$
The conditioning on $p$ enters \emph{only} through the averaging set $S_p$: candidate neighbors in $j^A(i)$ and target ranks in $r^B(i,j)$ are drawn from the full token set, not restricted to $p$. This design choice is deliberate, restricting the neighbor search itself to tokens of the same PoS would make $\Delta_p$ blind, by construction, to cross-category reattachment (e.g., a determiner's nearest neighbor shifting from another determiner to its head noun), which is precisely the phenomenon we aim to detect.

As with the unconditional imbalance, $\Delta_p \to 2/N$ indicates a fully preserved neighborhood between $A$ and $B$ for tokens of category $p$, while $\Delta_p \to 1$ indicates statistically independent neighborhoods; in general $\Delta_p(A\to B) \neq \Delta_p(B\to A)$, and this asymmetry, rather than the raw magnitude,  carries the relevant signal: $\Delta_p(A\to B) \approx 0$ indicates that $A$'s neighborhood structure for category $p$ is predictive of $B$'s, i.e.\ the information in $B$ is contained in $A$, not vice versa.

We compute $\Delta_p(\ell \to \ell_{\text{first}})$ and $\Delta_p(\ell \to \ell_{\text{last}})$ for every layer $\ell$ and every POS category $p$. A crossover between the two profiles, $\Delta_p(\ell\to\ell_{\text{first}})$ rising from the floor while $\Delta_p(\ell\to\ell_{\text{last}})$ falls to it, localizes the layer range at which tokens of category $p$ detach from their input-level identity and converge toward their final contextual role.

\section{Ablation Studies}

\begin{figure}
    \centering
    \includegraphics[width=1\linewidth]{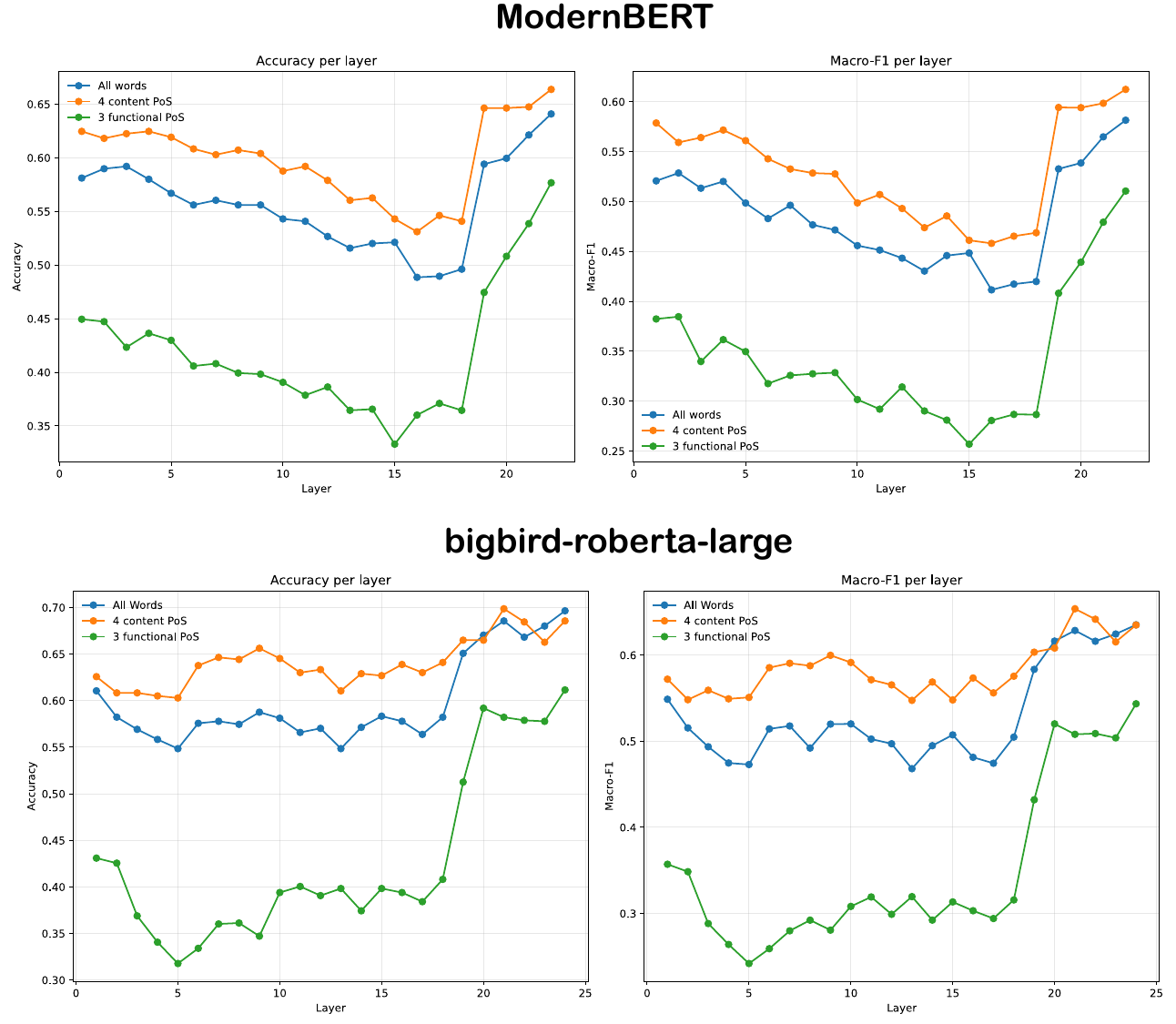}
    \caption{It shows the accuracy and F1 score of a logistic regression model fitted on the average of: all representations (\textcolor{cyan}{blue}), 4 content PoS (NOUN, VERB, ADJ, ADV) (\textcolor{orange}{orange}) and 3 function PoS (DET, ADP, PRON) (\textcolor{teal}{green}) for both \texttt{ModernBERT} and \texttt{bigbird-roberta-large}. }
    \label{acc1}
\end{figure}
\begin{table}[t]
\centering
\small
\begin{tabular}{lr}
\hline
\textbf{Specialty} & \textbf{Frequency} \\
\hline
Oncology & 144 \\
Neurology & 140 \\
Gastroenterology & 90 \\
Cardiology & 88 \\
Infectious Disease & 75 \\
Endocrinology & 60 \\
Psychiatry & 52 \\
Orthopedics & 45 \\
Hematology & 45 \\
Nephrology & 44 \\
Pulmonology & 44 \\
Gynecology/Obstetrics & 36 \\
Dermatology & 33 \\
Rheumatology & 17 \\
Pediatrics & 15 \\
\hline
\textbf{Total} & \textbf{928} \\
\hline
\end{tabular}
\caption{Distribution of the PubMed abstracts across the 15 medical specialties.}
\label{specialties}
\end{table}
\begin{figure*}
    \centering
    \includegraphics[width=1\linewidth]{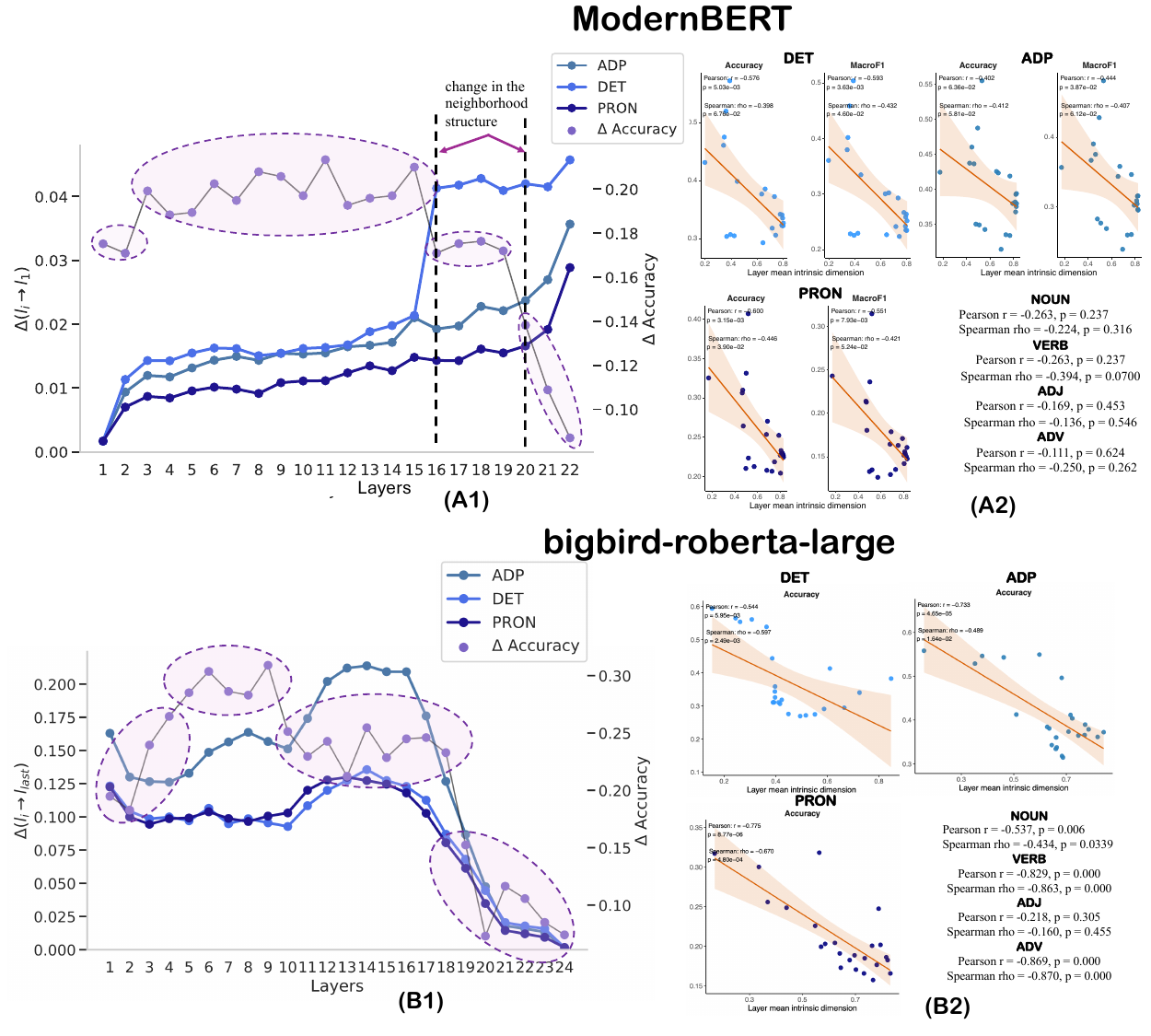}
    \caption{\textbf{(A)} shows the PoS-conditional Information Imbalance from layer $i$ to layer 1, together with the accuracy gap between content words and function words across layers in \texttt{ModernBERT} and \texttt{bigbird-roberta-large}. \textbf{(B)} shows the correlation between accuracy and cID for each PoS.}
    \label{acc2}
\end{figure*}

\subsection{Semantic Absorption in Function Words}
\label{classtest}

The analyses in the main text characterize how the geometry of a token's representation
evolves across layers, but they do not tell us what that reorganization is \emph{for}. To
address this, we introduce a downstream multiclass classification task designed to track
the emergence of semantic content in categories that carry none lexically.

\paragraph{Rationale.} The distinction between content and function words is, at its core,
a distinction in the source of meaning. Content words are open-class items whose semantic
contribution is lexically specified: a noun denotes independently of the sentence it
occurs in. Function words form a closed class whose contribution is relational, they
signal how other constituents combine, and in isolation they are close to semantically
vacuous. A determiner does not denote a domain of discourse; it binds a nominal. Under
this view, a representation averaged over the function words of a document should carry
little information about what that document is about, at least at the input layer. What
the attention mechanism makes possible, however, is that a function word progressively
absorbs the semantics of the constituents it governs, acquiring, through contextual
integration, information it never possessed lexically. If this is what happens, the
document-level predictivity of function words should be near-chance early in the network
and rise sharply at the layers where their neighborhood structure is reorganized.

\paragraph{Setup.} For each document we build three averaged representations, over all
tokens, over content words only (NOUN, VERB, ADJ, ADV), and over function words only
(DET, PRON, ADP), and use each as input to a logistic regression classifier (using sklearn's default hyperparameters), fitted
independently at every layer (using a Stratified-K-fold cross validation, with K = 5). We restrict this experiment to the two encoders. In a
decoder, a token's representation is conditioned only on its left context, so a function
word can absorb the semantics of its governor but not of the complement it introduces,
which in English typically follows it, an asymmetry that would confound the measurement
we are after.

\paragraph{Data.} We use the subset of \texttt{Pile-10k} consisting of PubMed abstracts,
which were labeled by \texttt{Claude Opus 4.8} (and later manually checked by an expert clinician) with one of 
15 medical specialties. This is a domain where the
discriminative burden falls almost entirely on nominal vocabulary: the specialty of an
abstract is signalled by its technical nouns, so content words provide a natural upper
bound and function words a natural floor. Labels statistics are reported in Table \ref{specialties}.

\paragraph{Results.} As shown in Figure~\ref{acc1}, in both \texttt{ModernBERT} and
\texttt{bigbird-roberta-large} content words are already highly discriminative in the earliest
layers, confirming that lexical semantics alone suffices for this task and dominates
contextual information. Around layer $19$ both models show a marked increase in
performance, considerably steeper for function words than for content words, to the point
that the two groups nearly converge. Function words, in other words, end up almost as
informative about the topic of a document as the nouns that encode it lexically, despite
starting from a far lower baseline.

To identify what drives this convergence, Figure~\ref{acc2} \textbf{(A1)} and \textbf{(B1)} plots the layer-wise
gap in accuracy between content and function words alongside the cII
profiles, computed with respect to the first layer for \texttt{ModernBERT} and to the last layer
for \texttt{bigbird-roberta-large}. In \texttt{ModernBERT}, the rise in cII for DET coincides with the drop
in the accuracy gap: precisely when determiners detach from their input-level
neighborhoods, they begin to behave like content words for the purposes of the task. In
\texttt{bigbird-roberta-large}, the cII profiles of all three function categories track the
accuracy gap directly. In both cases the same conclusion holds, semantic absorption is
not gradual but localized, and it coincides with an identifiable restructuring of the
local neighborhood, consistent with the interpretation that a function word acquires
meaning by attaching to the constituent it governs.

Figures~\ref{acc2} \textbf{(A2)} \textbf{(B2)} test the complementary relation with
ID, fitting the classifier on the averaged representation of a single
PoS at a time. We find a significantly negative correlation between layer-mean cID and
classification accuracy, strongest for DET, PRON and ADP: the layers at which a category
is most semantically informative are those at which its local manifold is most
compressed, suggesting that the resolution of a grammatical dependency and the collapse
of the representational degrees of freedom associated with it are two descriptions of the
same event.

\begin{figure*}
    \centering
    \includegraphics[width=1\linewidth]{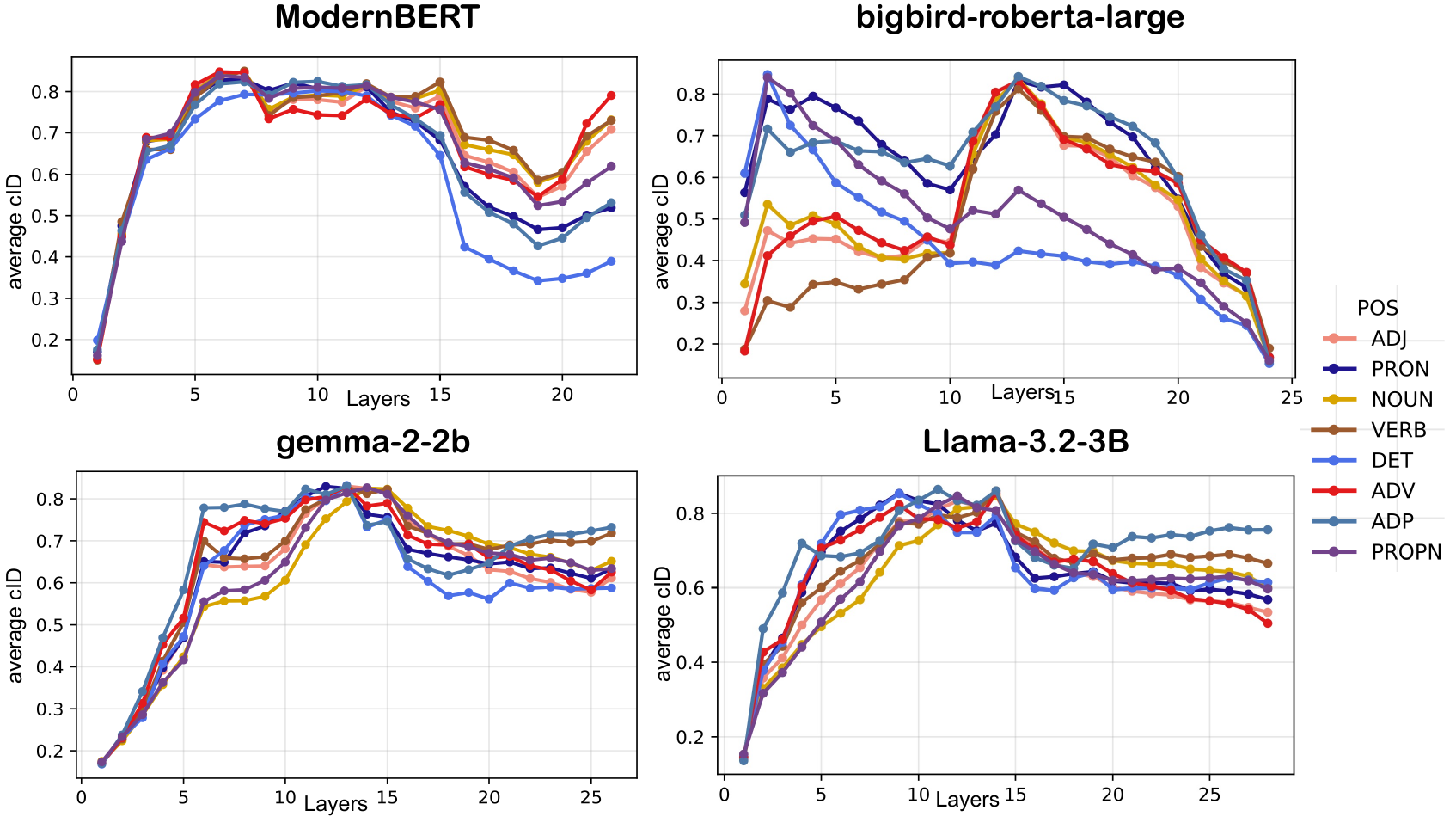}
    \caption{Normalized PoS-conditioned ID trajectories for the four language models considered.}
    \label{nonormalized}
\end{figure*}
\subsection{Normalized cID trajectories}
\label{nonormalized_section}

Because the absolute cID scale differs across PoS categories and models, comparing the
\emph{shape} of the trajectories requires removing this offset. For each category we
therefore rescale the layer-wise estimates $\mu_\ell$, with standard errors
$\sigma_\ell$, to the unit interval, using
\begin{equation}
y_{\text{shift}} = \min_\ell (\mu_\ell - \sigma_\ell),
\qquad
y_{\text{scale}} = \max_\ell (\mu_\ell + \sigma_\ell),
\end{equation}
so that
\begin{equation}
y_\ell = \frac{\mu_\ell - y_{\text{shift}}}{y_{\text{scale}} - y_{\text{shift}}},
\qquad
y_\ell^{\text{err}} = \frac{\sigma_\ell}{y_{\text{scale}} - y_{\text{shift}}}.
\end{equation}
The bounds are taken over the confidence band rather than the point estimates, so that
the full interval falls within $[0,1]$. Since the transformation is affine and applied
per category, it preserves the relative position of peaks and troughs across layers while
making trajectories with different baseline dimensionality directly comparable, which is
what Figure~\ref{nonormalized} shows.

\section{Point-wise Geometric Features}
\label{pointwise}

The measures introduced in the main text are group-level: cID and cII are averaged over
all tokens of a given PoS, which is what makes them informative about a category as a
whole but leaves them unusable as per-token features. The classification experiments
require the opposite granularity, one feature vector per token, and the cID results
themselves motivate this move: if intrinsic dimension varies systematically by PoS, then
a single global estimate is not the right descriptor for an individual token, whose
manifold is locally shaped by the category it belongs to. We therefore turn to a
\emph{local} estimator, one that characterizes the manifold in the immediate
neighborhood of each point rather than over the token set as a whole.

We use the Maximum Likelihood Estimator \citep{levina2004maximum} as implemented in
\texttt{scikit-dimension} \citep{bac2021scikit}, which models the distances from a point
to its $k$ nearest neighbors as a non-homogeneous Poisson process and derives the local
ID by maximizing the log-likelihood of the observed distances. Estimation is carried out
within the representation space of the token's own context, so that the resulting value
reflects the geometric complexity of the manifold surrounding that specific token, layer
by layer.

We obtain point-wise cII analogously. Rather than averaging neighbor ranks over a PoS
category, we retain the exact rank of each individual token, recording
$\Delta_p(\ell \to \ell_{\text{first}})$ and $\Delta_p(\ell \to \ell_{\text{last}})$ at
every layer. The resulting per-token ID and II trajectories are then used as feature
vectors in the classification experiments.
Tables~\ref{binary_comparison} and~\ref{pos_comparison} report per-model results
alongside those obtained by aggregating features across the four models. 
Tables~\ref{model_comparison_binary_propn} and~\ref{model_comparison_multilabel_propn}
report the results obtained by including PROPN among content words (initially
excluded from the classification experiments because of its distinctive behavior), where performance
only slightly degrades across models and tasks.

\begin{table*}[t]
\centering
\tiny
\setlength{\tabcolsep}{4pt}

\begin{minipage}[t]{0.48\textwidth}
\centering
\begin{tabular}{lccccc}
\hline
\textbf{Method} & \textbf{Accuracy} & \textbf{Precision} & \textbf{Recall} & \textbf{F1} & \textbf{ROC-AUC} \\
\hline
\multicolumn{6}{c}{\cellcolor{mygray} \texttt{ModernBERT}} \\
\hline
ID + II & \val{0.814}{0.009} & \val{0.805}{0.013} & \val{0.657}{0.022} & \val{0.723}{0.016} & \val{0.871}{0.009} \\ \hline
II & \val{0.785}{0.008} & \val{0.808}{0.017} & \val{0.549}{0.021} & \val{0.654}{0.017} & \val{0.836}{0.010} \\ \hline
ID & \val{0.770}{0.009} & \val{0.697}{0.013} & \val{0.670}{0.022} & \val{0.683}{0.016} & \val{0.831}{0.010} \\ \hline
Shuffled & \val{0.667}{0.009} & \val{0.559}{0.016} & \val{0.481}{0.028} & \val{0.517}{0.021} & \val{0.712}{0.010} \\ \hline
\multicolumn{6}{c}{\cellcolor{mygray} \texttt{bigbird-roberta-large}} \\ \hline
ID + II & \val{0.811}{0.009} & \val{0.776}{0.020} & \val{0.691}{0.017} & \val{0.731}{0.012} & \val{0.874}{0.004} \\ \hline
II & \val{0.676}{0.005} & \val{0.673}{0.028} & \val{0.248}{0.015} & \val{0.362}{0.018} & \val{0.670}{0.012} \\ \hline
ID & \val{0.790}{0.009} & \val{0.746}{0.018} & \val{0.660}{0.023} & \val{0.700}{0.014} & \val{0.852}{0.006} \\ \hline
Shuffled & \val{0.615}{0.009} & \val{0.473}{0.022} & \val{0.337}{0.035} & \val{0.393}{0.030} & \val{0.669}{0.012} \\ \hline
\multicolumn{6}{c}{\cellcolor{mygray} \texttt{Gemma-2-2B}} \\ \hline
ID + II & \val{0.799}{0.007} & \val{0.737}{0.011} & \val{0.712}{0.019} & \val{0.725}{0.013} & \val{0.864}{0.007} \\ \hline
II & \val{0.603}{0.012} & \val{0.449}{0.025} & \val{0.306}{0.027} & \val{0.363}{0.023} & \val{0.668}{0.014} \\ \hline
ID & \val{0.802}{0.008} & \val{0.778}{0.011} & \val{0.654}{0.029} & \val{0.710}{0.018} & \val{0.863}{0.007} \\ \hline
Shuffled & \val{0.616}{0.012} & \val{0.474}{0.025} & \val{0.323}{0.025} & \val{0.384}{0.025} & \val{0.669}{0.013} \\ \hline
\multicolumn{6}{c}{\cellcolor{mygray} \texttt{Llama-3.2-3B}} \\ \hline
ID + II & \val{0.804}{0.006} & \val{0.789}{0.012} & \val{0.646}{0.017} & \val{0.710}{0.013} & \val{0.857}{0.008} \\ \hline
II & \val{0.726}{0.010} & \val{0.706}{0.023} & \val{0.449}{0.018} & \val{0.549}{0.017} & \val{0.759}{0.012} \\ \hline
ID & \val{0.768}{0.009} & \val{0.760}{0.019} & \val{0.547}{0.027} & \val{0.636}{0.020} & \val{0.813}{0.011} \\ \hline
Shuffled & \val{0.608}{0.008} & \val{0.374}{0.040} & \val{0.087}{0.022} & \val{0.140}{0.031} & \val{0.592}{0.015} \\ \hline
\multicolumn{6}{c}{\cellcolor{mygray} \texttt{All models}} \\ \hline

        ID + II & \val{0.878}{0.008} & \val{0.861}{0.012} & \val{0.802}{0.017} & \val{0.830}{0.012} & \val{0.937}{0.005} \\ \hline
        II & \val{0.830}{0.007} & \val{0.836}{0.014} & \val{0.677}{0.016} & \val{0.748}{0.012} & \val{0.886}{0.006} \\ \hline
        ID & \val{0.850}{0.007} & \val{0.815}{0.014} & \val{0.773}{0.015} & \val{0.793}{0.009} & \val{0.907}{0.005} \\ \hline
        Shuffled & \val{0.593}{0.013} & \val{0.441}{0.019} & \val{0.349}{0.025} & \val{0.389}{0.021} & \val{0.618}{0.014} \\ \hline
\end{tabular}
\caption{Binary classification (\textbf{function} words vs.\ \textbf{content} words) performances for \texttt{ModernBERT}, \texttt{bigbird-roberta-large}, \texttt{gemma-2-2B}, \texttt{Llama-3.2-3B}, and all models feature combined, averaged over 20 runs.}
\label{binary_comparison}
\end{minipage}
\hfill
\begin{minipage}[t]{0.48\textwidth}
\centering
\begin{tabular}{lccccc}
\hline
\textbf{Method} & \textbf{Accuracy} & \textbf{Precision} & \textbf{Recall} & \textbf{F1} & \textbf{ROC-AUC} \\
\hline
\multicolumn{6}{c}{\cellcolor{mygray} \texttt{ModernBERT}} \\
\hline
ID + II & \val{0.394}{0.011} & \val{0.453}{0.010} & \val{0.394}{0.011} & \val{0.399}{0.011} & \val{0.768}{0.005} \\ \hline
II & \val{0.347}{0.007} & \val{0.415}{0.011} & \val{0.347}{0.007} & \val{0.337}{0.009} & \val{0.729}{0.006} \\ \hline
ID & \val{0.323}{0.014} & \val{0.364}{0.015} & \val{0.323}{0.014} & \val{0.319}{0.015} & \val{0.724}{0.006} \\ \hline
Shuffled & \val{0.183}{0.007} & \val{0.235}{0.014} & \val{0.183}{0.007} & \val{0.165}{0.008} & \val{0.599}{0.008} \\ \hline
\multicolumn{6}{c}{\cellcolor{mygray} \texttt{bigbird-roberta-large}} \\ \hline
ID + II & \val{0.358}{0.011} & \val{0.401}{0.014} & \val{0.358}{0.011} & \val{0.361}{0.011} & \val{0.754}{0.007} \\ \hline
II & \val{0.202}{0.013} & \val{0.274}{0.018} & \val{0.202}{0.013} & \val{0.203}{0.016} & \val{0.622}{0.006} \\ \hline
ID & \val{0.331}{0.011} & \val{0.375}{0.013} & \val{0.331}{0.011} & \val{0.326}{0.011} & \val{0.742}{0.005} \\ \hline
Shuffled & \val{0.189}{0.009} & \val{0.220}{0.018} & \val{0.189}{0.009} & \val{0.168}{0.010} & \val{0.574}{0.007} \\ \hline
\multicolumn{6}{c}{\cellcolor{mygray} \texttt{Gemma-2-2B}} \\ \hline
ID + II & \val{0.412}{0.010} & \val{0.466}{0.010} & \val{0.412}{0.010} & \val{0.416}{0.010} & \val{0.774}{0.005} \\ \hline
II & \val{0.283}{0.008} & \val{0.354}{0.011} & \val{0.283}{0.008} & \val{0.284}{0.009} & \val{0.659}{0.006} \\ \hline
ID & \val{0.370}{0.011} & \val{0.422}{0.012} & \val{0.370}{0.011} & \val{0.371}{0.011} & \val{0.757}{0.006} \\ \hline
Shuffled & \val{0.182}{0.008} & \val{0.216}{0.013} & \val{0.182}{0.008} & \val{0.162}{0.008} & \val{0.574}{0.008} \\ \hline
\multicolumn{6}{c}{\cellcolor{mygray} \texttt{Llama-3.2-3B}} \\ \hline
ID + II & \val{0.391}{0.012} & \val{0.440}{0.011} & \val{0.391}{0.012} & \val{0.400}{0.012} & \val{0.765}{0.007} \\ \hline
II & \val{0.311}{0.011} & \val{0.367}{0.010} & \val{0.311}{0.011} & \val{0.321}{0.010} & \val{0.697}{0.006} \\ \hline
ID & \val{0.325}{0.013} & \val{0.377}{0.013} & \val{0.325}{0.013} & \val{0.325}{0.014} & \val{0.722}{0.009} \\ \hline
Shuffled & \val{0.158}{0.009} & \val{0.202}{0.013} & \val{0.158}{0.009} & \val{0.141}{0.009} & \val{0.535}{0.007} \\ \hline
\multicolumn{6}{c}{\cellcolor{mygray} \texttt{All models}} \\ \hline
ID + II & \val{0.529}{0.010} & \val{0.567}{0.010} & \val{0.529}{0.010} & \val{0.540}{0.010} & \val{0.844}{0.005} \\ \hline
        II & \val{0.424}{0.008} & \val{0.486}{0.009} & \val{0.424}{0.008} & \val{0.441}{0.009} & \val{0.789}{0.006} \\ \hline
        ID & \val{0.486}{0.012} & \val{0.512}{0.013} & \val{0.486}{0.012} & \val{0.487}{0.013} & \val{0.816}{0.006} \\ \hline
        Shuffled & \val{0.171}{0.006} & \val{0.207}{0.011} & \val{0.171}{0.006} & \val{0.167}{0.007} & \val{0.534}{0.007} \\ \hline
\end{tabular}
\caption{PoS classification performances for \texttt{ModernBERT}, \texttt{bigbird-roberta-large}, \texttt{gemma-2-2B}, \texttt{Llama-3.2-3B}, and all models feature combined, averaged over 20 runs.}    \label{model_comparison}
\label{pos_comparison}
\end{minipage}

\end{table*}

\begin{table*}[t]
\centering
\tiny
\setlength{\tabcolsep}{3pt}

\begin{minipage}[t]{0.48\textwidth}
\centering
\begin{tabular}{lccccc}
\hline
\textbf{Method} & \textbf{Acc.} & \textbf{Prec.} & \textbf{Rec.} & \textbf{F1} & \textbf{ROC-AUC} \\
\hline
\multicolumn{6}{c}{\cellcolor{mygray} \texttt{ModernBERT}} \\ \hline
ID + II & \val{0.830}{0.010} & \val{0.809}{0.019} & \val{0.618}{0.023} & \val{0.700}{0.019} & \val{0.868}{0.011} \\ \hline
II & \val{0.801}{0.009} & \val{0.801}{0.020} & \val{0.508}{0.021} & \val{0.621}{0.019} & \val{0.832}{0.012} \\ \hline
ID & \val{0.796}{0.014} & \val{0.765}{0.038} & \val{0.530}{0.028} & \val{0.625}{0.026} & \val{0.820}{0.013} \\ \hline
Shuffled & \val{0.633}{0.008} & \val{0.257}{0.037} & \val{0.078}{0.024} & \val{0.119}{0.031} & \val{0.671}{0.011} \\ \hline
\multicolumn{6}{c}{\cellcolor{mygray} \texttt{bigbird-roberta-large}} \\ \hline
ID + II & \val{0.810}{0.007} & \val{0.792}{0.017} & \val{0.548}{0.020} & \val{0.648}{0.016} & \val{0.853}{0.007} \\ \hline
II & \val{0.709}{0.006} & \val{0.644}{0.030} & \val{0.201}{0.017} & \val{0.306}{0.021} & \val{0.681}{0.012} \\ \hline
ID & \val{0.785}{0.010} & \val{0.760}{0.023} & \val{0.478}{0.027} & \val{0.586}{0.023} & \val{0.830}{0.011} \\ \hline
Shuffled & \val{0.663}{0.011} & \val{0.167}{0.051} & \val{0.013}{0.007} & \val{0.024}{0.011} & \val{0.619}{0.012} \\ \hline
\multicolumn{6}{c}{\cellcolor{mygray} \texttt{Gemma-2-2B}} \\ \hline
ID + II & \val{0.803}{0.009} & \val{0.754}{0.017} & \val{0.567}{0.021} & \val{0.647}{0.017} & \val{0.858}{0.007} \\ \hline
II & \val{0.674}{0.008} & \val{0.445}{0.041} & \val{0.087}{0.018} & \val{0.145}{0.026} & \val{0.650}{0.016} \\ \hline
ID & \val{0.796}{0.006} & \val{0.783}{0.014} & \val{0.499}{0.021} & \val{0.609}{0.017} & \val{0.856}{0.007} \\ \hline
Shuffled & \val{0.656}{0.009} & \val{0.226}{0.050} & \val{0.035}{0.015} & \val{0.059}{0.023} & \val{0.632}{0.013} \\ \hline
\multicolumn{6}{c}{\cellcolor{mygray} \texttt{Llama-3.2-3B}} \\ \hline
ID + II & \val{0.814}{0.008} & \val{0.780}{0.018} & \val{0.588}{0.024} & \val{0.670}{0.017} & \val{0.852}{0.010} \\ \hline
II & \val{0.748}{0.008} & \val{0.690}{0.022} & \val{0.389}{0.020} & \val{0.497}{0.020} & \val{0.764}{0.008} \\ \hline
ID & \val{0.758}{0.017} & \val{0.773}{0.030} & \val{0.352}{0.075} & \val{0.478}{0.074} & \val{0.796}{0.020} \\ \hline
Shuffled & \val{0.679}{0.004} & \val{0.100}{0.300} & \val{0.000}{0.000} & \val{0.000}{0.001} & \val{0.552}{0.012} \\ \hline
\end{tabular}
\caption{Binary classification (\textbf{function} vs.\ \textbf{content} words) including PROPN, averaged over 20 runs.}
\label{model_comparison_binary_propn}
\end{minipage}
\hfill
\begin{minipage}[t]{0.48\textwidth}
\centering
\begin{tabular}{lccccc}
\hline
\textbf{Method} & \textbf{Acc.} & \textbf{Prec.} & \textbf{Rec.} & \textbf{F1} & \textbf{ROC-AUC} \\
\hline
\multicolumn{6}{c}{\cellcolor{mygray} \texttt{ModernBERT}} \\ \hline
ID + II & \val{0.362}{0.008} & \val{0.403}{0.012} & \val{0.362}{0.008} & \val{0.351}{0.009} & \val{0.754}{0.005} \\ \hline
II & \val{0.301}{0.011} & \val{0.362}{0.015} & \val{0.301}{0.011} & \val{0.282}{0.010} & \val{0.710}{0.006} \\ \hline
ID & \val{0.288}{0.009} & \val{0.325}{0.015} & \val{0.288}{0.009} & \val{0.277}{0.010} & \val{0.717}{0.005} \\ \hline
Shuffled & \val{0.155}{0.006} & \val{0.192}{0.011} & \val{0.155}{0.006} & \val{0.138}{0.006} & \val{0.590}{0.006} \\ \hline
\multicolumn{6}{c}{\cellcolor{mygray} \texttt{bigbird-roberta-large}} \\ \hline
ID + II & \val{0.360}{0.010} & \val{0.391}{0.015} & \val{0.360}{0.010} & \val{0.357}{0.012} & \val{0.764}{0.006} \\ \hline
II & \val{0.230}{0.009} & \val{0.255}{0.015} & \val{0.230}{0.009} & \val{0.203}{0.010} & \val{0.644}{0.008} \\ \hline
ID & \val{0.327}{0.010} & \val{0.363}{0.012} & \val{0.327}{0.010} & \val{0.320}{0.010} & \val{0.745}{0.006} \\ \hline
Shuffled & \val{0.169}{0.009} & \val{0.201}{0.012} & \val{0.169}{0.009} & \val{0.157}{0.010} & \val{0.594}{0.008} \\ \hline
\multicolumn{6}{c}{\cellcolor{mygray} \texttt{Gemma-2-2B}} \\ \hline
ID + II & \val{0.384}{0.007} & \val{0.428}{0.009} & \val{0.384}{0.007} & \val{0.382}{0.008} & \val{0.770}{0.006} \\ \hline
II & \val{0.261}{0.008} & \val{0.318}{0.010} & \val{0.261}{0.008} & \val{0.259}{0.008} & \val{0.666}{0.005} \\ \hline
ID & \val{0.344}{0.007} & \val{0.388}{0.009} & \val{0.344}{0.007} & \val{0.340}{0.008} & \val{0.750}{0.005} \\ \hline
Shuffled & \val{0.153}{0.009} & \val{0.188}{0.010} & \val{0.153}{0.009} & \val{0.137}{0.008} & \val{0.572}{0.007} \\ \hline
\multicolumn{6}{c}{\cellcolor{mygray} \texttt{Llama-3.2-3B}} \\ \hline
ID + II & \val{0.352}{0.014} & \val{0.392}{0.014} & \val{0.352}{0.014} & \val{0.348}{0.014} & \val{0.754}{0.009} \\ \hline
II & \val{0.294}{0.010} & \val{0.329}{0.012} & \val{0.294}{0.010} & \val{0.288}{0.011} & \val{0.689}{0.007} \\ \hline
ID & \val{0.287}{0.018} & \val{0.330}{0.018} & \val{0.287}{0.018} & \val{0.277}{0.019} & \val{0.715}{0.011} \\ \hline
Shuffled & \val{0.141}{0.007} & \val{0.176}{0.012} & \val{0.141}{0.007} & \val{0.132}{0.007} & \val{0.553}{0.006} \\ \hline
\end{tabular}
\caption{PoS classification including PROPN, averaged over 20 runs.}
\label{model_comparison_multilabel_propn}
\end{minipage}

\end{table*}

\end{document}